\documentclass[twocolumn]{article}  
\usepackage{graphicx}
\usepackage[T1]{fontenc}
\usepackage[subrefformat=parens]{subcaption}

\makeatletter
\let\cl@chapter\undefined
\makeatletter

\usepackage{amsmath,amssymb}   
\usepackage{amsthm} 
\usepackage{mathtools} 
\usepackage[numbers,sort&compress]{natbib}
\usepackage{hyperref}
\hypersetup{colorlinks=true,pdfborder={0 0 0}}
\usepackage{cleveref}  
\usepackage{amsbsy}
\usepackage{bm}
\usepackage{url}
\usepackage{color}
\usepackage{multirow}
\usepackage{booktabs}
\usepackage{autobreak}
\allowdisplaybreaks 

\newcommand{\inlineTag}{
    \refstepcounter{equation}
    \bgroup\normalfont\normalcolor (\theequation)\egroup}

\crefname{equation}{Eq.}{Eqs.}%
\crefname{figure}{Fig.}{Figs.}%

\usepackage{algorithm}
\usepackage[noend]{algpseudocode}

\algnewcommand{\Break}{\textbf{break}}
\algrenewcommand\algorithmicindent{1em}

\everydisplay{\small} 

\newcommand\Erase{\bgroup\markoverwith{\textcolor{red}{\rule[.5ex]{2pt}{0.4pt}}}\ULon}

\begin{document}

    \title{
        Conceptual Design on the Field of View of Celestial Navigation Systems for Maritime Autonomous Surface Ships
    }
    
    \author{
        Kouki Wakita$^{1}$ \and
        Fuyuki Hane$^{2}$ \and
        Takeshi Sekiguchi$^{3}$ \and
        Shigehito Shimizu$^{3}$ \and
        Shinji Mitani$^{3}$ \and
        Youhei Akimoto$^{4,5}$ \and
        Atsuo Maki$^{1}$
    }
    
    \date{%
    \flushleft{\footnotesize
        $^1$Osaka University, 2-1 Yamadaoka, Suita, Osaka, Japan \\%
        $^2$TOKYO KEIKI INC., 333-4, Azuma, Yaita, Tochigi, Japan \\%
        $^3$Japan Aerospace Exploration Agency (JAXA), 2-1-1 Sengen, Tsukuba, Ibaraki, Japan \\%
        $^4$University of Tsukuba, 1-1-1 Tennodai, Tsukuba, Ibaraki, Japan \\
        $^5$RIKEN Center for Advanced Intelligence Project, 1-4-1 Nihonbashi, Chuo-ku, Tokyo, Japan \\[2ex]%
        Email: kouki\_wakita@naoe.eng.osaka-u.ac.jp; maki@naoe.eng.osaka-u.ac.jp \\
    }}


    \maketitle

    \begin{abstract}
        In order to understand the appropriate field of view (FOV) size of celestial automatic navigation systems for surface ships, we investigate the variations of measurement accuracy of star position and probability of successful star identification with respect to FOV, focusing on the decreasing number of observable star magnitudes and the presence of physically covered stars in marine environments. 
        The results revealed that, although a larger FOV reduces the measurement accuracy of star positions, it increases the number of observable objects and thus improves the probability of star identification using subgraph isomorphism-based methods. It was also found that, although at least four objects need to be observed for accurate identification, four objects may not be sufficient for wider FOVs. On the other hand, from the point of view of celestial navigation systems, a decrease in the measurement accuracy leads to a decrease in positioning accuracy. Therefore, it was found that maximizing the FOV is required for celestial automatic navigation systems as long as the desired positioning accuracy can be ensured. Furthermore, it was found that algorithms incorporating more than four observed celestial objects are required to achieve highly accurate star identification over a wider FOV.
        
    \end{abstract}

    \section{Introduction}\label{sec:intro}
        The measurement of the ship's position is essential for ship operations, and having a redundant positioning system is important for safe operation. In recent years, research and development of Maritime Autonomous Surface Ships (MASS) has been active, and the demand for redundant positioning systems for the realisation of MASS is increasing. The Global Navigation Satellite System (GNSS) is primarily used to determine the position of ships at sea. However, GNSS is an external reference system, that relies on signals from artificial satellites, and is susceptible to radio interference. The need for alternative positioning methods to GNSS has been pointed out by Kaplan \cite{Kaplan1999}. Therefore, ships are required to have positioning methods that can be used globally and are independent of any external system.

        One of the independent positioning methods is celestial navigation \cite{hasegawa1994}. Celestial navigation has been used by navigators to determine their ship's position before the spread of radio navigation technology. In celestial navigation, the accurate time is measured using a chronometer, and the altitude angles of celestial bodies such as the sun, moon, stars, and planets are measured using a sextant. Navigators also perform the identification of celestial bodies, called star identification (star ID), simultaneously with the measurement of celestial bodies. Based on the measured time and altitude angle, the latitude and longitude of the ship on the earth can be determined. The line of position (LOP) is often used to determine the ship's position, and it is calculated using the Nautical Almanac, which provides the positions of celestial bodies, and the Sight Reduction Tables, which assist in the calculation of the ship's position.
        
        In celestial navigation, it is necessary to measure the altitude of celestial bodies, identify celestial bodies, and calculate the ship's position. For the automation of celestial navigation for MASS, it is necessary to automate these processes. In the field of aerospace, research and development on automatic celestial navigation for attitude determination of spacecraft and aircraft have been conducted, and devices called Star Trackers are installed on spacecraft and are in practical use. In Star Trackers, stars are observed using image sensors, and the relative directions of stars are measured from the images. Several studies have been published on the calibration of image sensors \cite{Klaus2004,Zhang2017,Enright2018} and the centroiding techniques that achieves sub-pixel level measurement accuracy \cite{Samaan2002,Liebe2002}. Moreover, a number of efficient star ID methods have also been developed. Although the process of attitude determination is different, the techniques of measurement and star ID can be applied to the automation of celestial navigation for ships.

        However, the shipboard environment differs from the space environment, and previous research and development have mainly focused on the space environment. For example, spacecraft must be composed of radiation-resistant equipment and be as lightweight and unbreakable as possible. As a result, there are strict limitations on computational resources and storage capacity, requiring efficient star ID methods. Therefore, in addition to research to improve the accuracy of star ID, research on improving matching efficiency \cite{Mortari1997,Mortari2014,Zhao2016,Wang2018} and avoiding false stars \cite{Mueller2016} has been conducted.
        
\begin{figure}[t]
            \centering
            \includegraphics[width=0.8\hsize]{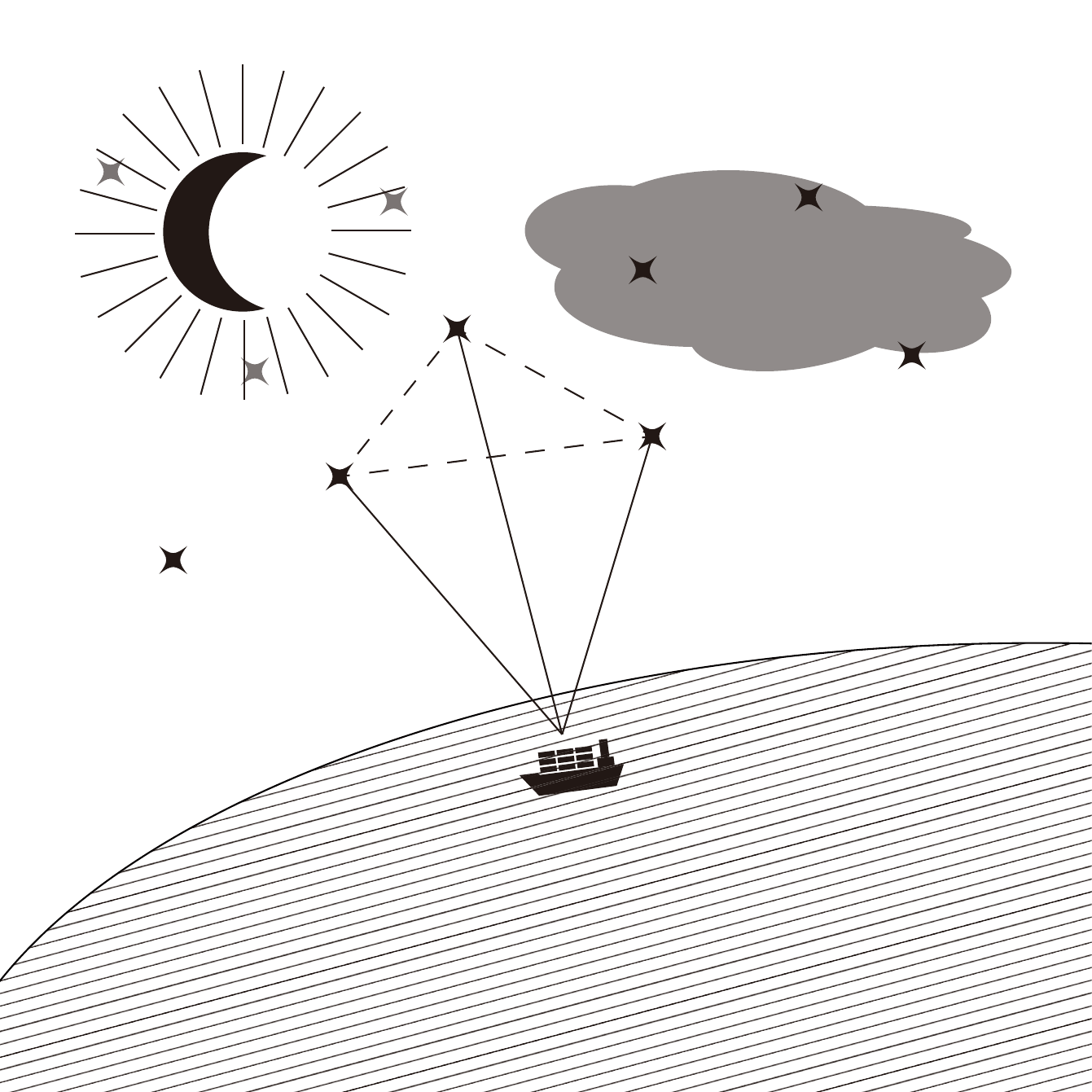}
            \caption{Celestial navigation on board autonomous ships}
            \label{fig:celestial_navigation}
        \end{figure}

        In contrast, in the shipboard environment, the effects of cosmic rays are almost negligible, and there are almost no restrictions on the weight of the equipment. However, the presence of an atmosphere makes the object relatively faint. Furthermore, the ship's motion may shorten the exposure time of the image sensor, reducing the number of stars that can be observed. Moreover, as shown in \Cref{fig:celestial_navigation}, the number of observable stars may also be reduced due to stars being covered by clouds or the moon. Keeping the number of observable stars high is important because the number of observable stars affects the accuracy of star ID.
    
        The simplest way to keep the number of observable stars high is to have a wide field of view (FOV). In Star Trackers designed for spacecraft, the FOV is at most $20^{\circ} \times 20^{\circ}$. For example, the FOV of the Next-generation Star Tracker (NSTT) is $16^{\circ} \times 16^{\circ}$ \cite{Sekiguchi2004}, the FOV of the ASTRO-APS Star Tracker is $20^{\circ} \times 20^{\circ}$, and the FOV of the High Accuracy Star Tracker (HAST) is $8.0^{\circ} \times 8.0^{\circ}$. Therefore, most studies do not focus on situations with an FOV larger than $20^{\circ} \times 20^{\circ}$. In the shipboard environment, where the number of observable stars may be reduced, it may be more appropriate to have a larger FOV than Star Trackers.

        Therefore, this paper discusses the required performance specifications for an automatic celestial navigation system in the shipboard environment from the perspective of the size of the FOV. Specifically, we investigate the variations of measurement accuracy of star position and probability of successful star ID with respect to the size of the FOV, focusing on the decrease in the magnitude of observable stars and the existence of physically covered stars in the shipboard environment. In particular, we assume the use of CMOS image sensors, which are also used in general cameras, for the measurement of stars, and calculate the maximum angular resolution for each FOV and the value of the angular resolution converted to sea distance. For the investigation of the probability of correct matching, we perform Monte Carlo simulations of star ID and obtain the probability of correct matching corresponding to each FOV.
    
        The remainder of this paper is structured as follows. \Cref{sec:note} shows the notation of the formulas used in this paper, \Cref{sec:measure} shows the results related to measurement accuracy,\Cref{sec:starid} shows the results related to the probability of successful star ID, and \Cref{sec:discuss} discusses the obtained results. Finally, \Cref{sec:conclude} presents the conclusions.   
    
    \section{Notation}\label{sec:note}

        In this section, we define the mathematical notations used throughout the paper. The set of real numbers is denoted by $\mathbb{R}$, and the $n$-dimensional Euclidean space is represented by $\mathbb{R}^{n}$. The set of angles is denoted by $\mathbb{S} = [0, 2\pi]$. The 3D rotation group is referred to as $SO(3)$.

        For a real number $a \in \mathbb{R}$, the sign function $\mathrm{sign}(a)$ returns 1 if $a > 0$, 0 if $a = 0$, and -1 if $a < 0$. When denoting two sets by $\mathcal{A}$ and $\mathcal{B}$, $\left|\mathcal{A}\right|$ represents the number of elements in $\mathcal{A}$. $\left\{a_{\in \mathcal{A}} \mid A \land B\right\}$ represents the subset of $\mathcal{A}$ whose elements satisfy both conditions $A$ and $B$. The empty set is denoted by $\emptyset$. $\mathcal{A} \times \mathcal{B}=\{(a, b) \mid a \in \mathcal{A} \wedge b \in \mathcal{B}\}$ represents the Cartesian product of $\mathcal{A}$ and $\mathcal{B}$. Finally, $\binom{n}{i} = \frac{n!}{i!(n-i)!}$ represents the number of ways to choose $i$ elements from $n$ elements.
        
    \section{Influence of FOV on measurements}\label{sec:measure}
        In this section, we present an estimate of the angular resolution with respect to the size of the FOV in order to discuss the measurement accuracy of stars. As previously mentioned, Star Trackers measure the positions of stars using image sensors. It is practical and economical to use image sensors for measuring the positions of stars at sea. Therefore, this study assumes the use of a general-purpose camera equipped with a CMOS image sensor. To obtain a simple approximation, we model the relationship between the star position and its projected position on the image sensor using a pinhole camera model.

        \subsection{Camera coordinates}\label{subsec:camera_coo}
            \begin{figure}[t]
                \centering
                \includegraphics[width=0.8\hsize]{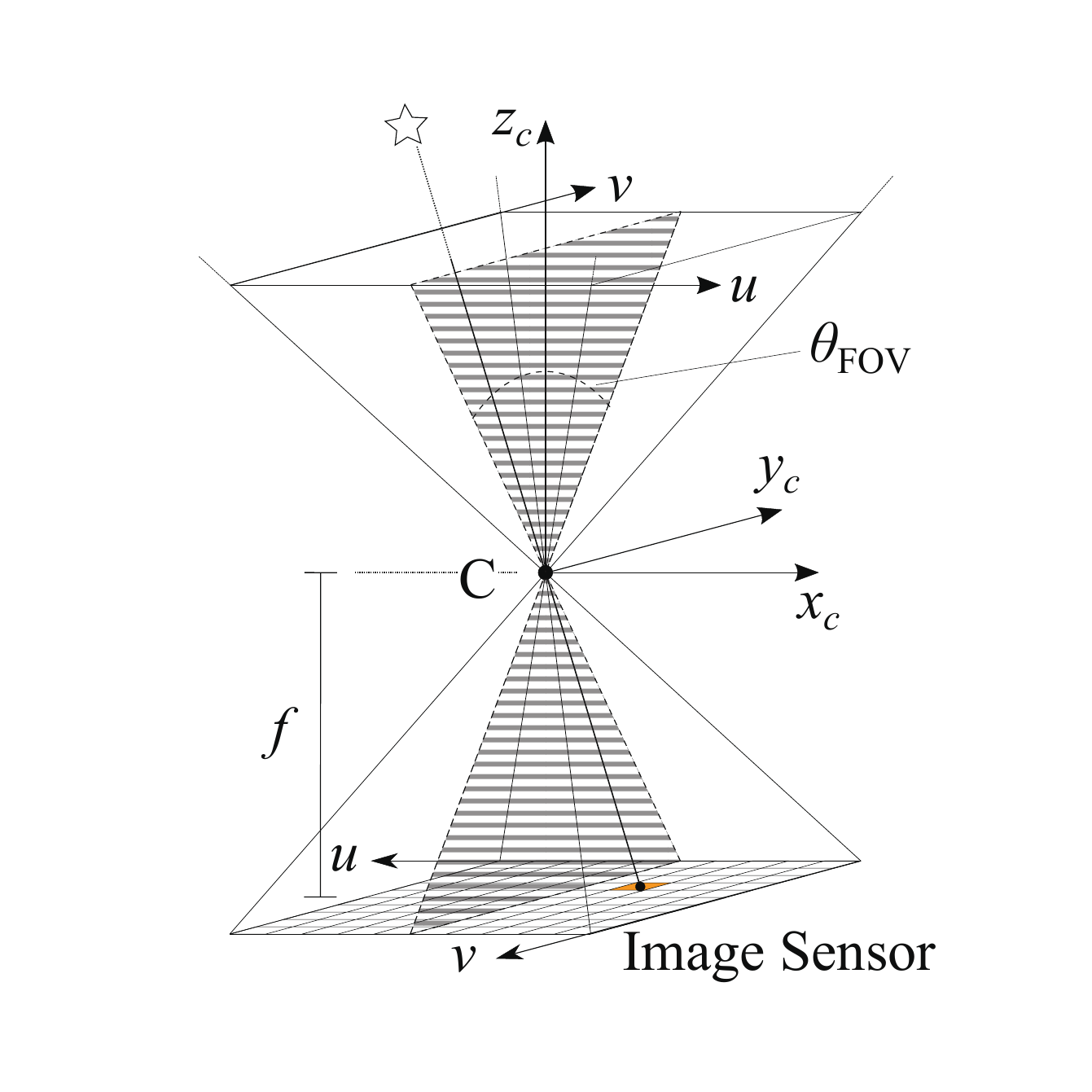}
                \caption{Camera coordinate systems}
                \label{fig:camera}
            \end{figure}
        
            First, we describe the coordinate system based on the measuring equipment. In this paper, we introduce a right-hand 3D orthogonal coordinate system $\mathrm{C}$-$x_{\mathrm{c}}y_{\mathrm{c}}z_{\mathrm{c}}$, with the focal point of the camera as the origin. As shown in \Cref{fig:camera}, we consider an image sensor with a focal length $f$ and the number of a pixel array $U \times U$. The measured stars are detected as light points on the image sensor, and their positions are measured as pixel coordinates. Using the pinhole camera model, when a star located at $(x_{\mathrm{c}}, y_{\mathrm{c}}, z_{\mathrm{c}})$ in the $\mathrm{C}$-$x_{\mathrm{c}}y_{\mathrm{c}}z_{\mathrm{c}}$ coordinate system is captured, its pixel coordinates $(u, v)$ are represented as follows:
            \begin{equation}
                s \begin{pmatrix}
                    u \\
                    v \\
                    1
                \end{pmatrix}=\begin{bmatrix}
                    f_x & 0 & c_x \\
                    0 & f_y & c_y \\
                    0 & 0 & 1
                \end{bmatrix}\begin{pmatrix}
                    x_{\mathrm{c}} \\
                    y_{\mathrm{c}} \\
                    z_{\mathrm{c}} 
                \end{pmatrix}
                \label{eq:C2pixel}
            \end{equation}
            Here, $s$ is a scale parameter, $f_x$ and $f_y$ are the focal lengths, and $c_x$ and $c_y$ are the principal points. The focal lengths $f_x, f_y$ and principal points $c_x, c_y$ are intrinsic camera parameters.
            By eliminating the scale parameter $s$ from \Cref{eq:C2pixel}, the position of a star in the $\mathrm{C}$-$x_{\mathrm{c}}y_{\mathrm{c}}z_{\mathrm{c}}$ coordinate system is expressed as follows:
            \begin{equation}
                \boldsymbol{s}_{\mathrm{c}} = \frac{1}{\sqrt{\tilde{u}^2+\tilde{v}^2+1}}\left(\begin{array}{c}
                    \tilde{u} \\
                    \tilde{v} \\
                    1 
                \end{array}\right)
                \label{eq:pixel}
            \end{equation}
            where $\tilde{u} = (u - c_x)/f_x$ and $\tilde{v} = (v - c_y)/f_y$.

            Therefore, given the camera parameters, it is possible to measure the position of stars in the $\mathrm{C}$-$x_{\mathrm{c}}y_{\mathrm{c}}z_{\mathrm{c}}$ coordinate system.
            
        \subsection{Approximate angular resolution}\label{subsec:precision}
            In this subsection, we present an estimate of the angular resolution per pixel based on the size of the FOV using a pinhole camera model. Here, we assume that the principal point is at the center of the image sensor and that lens distortion is not considered.
    
            \begin{figure}[t]
                \centering
                \includegraphics[width=0.8\hsize]{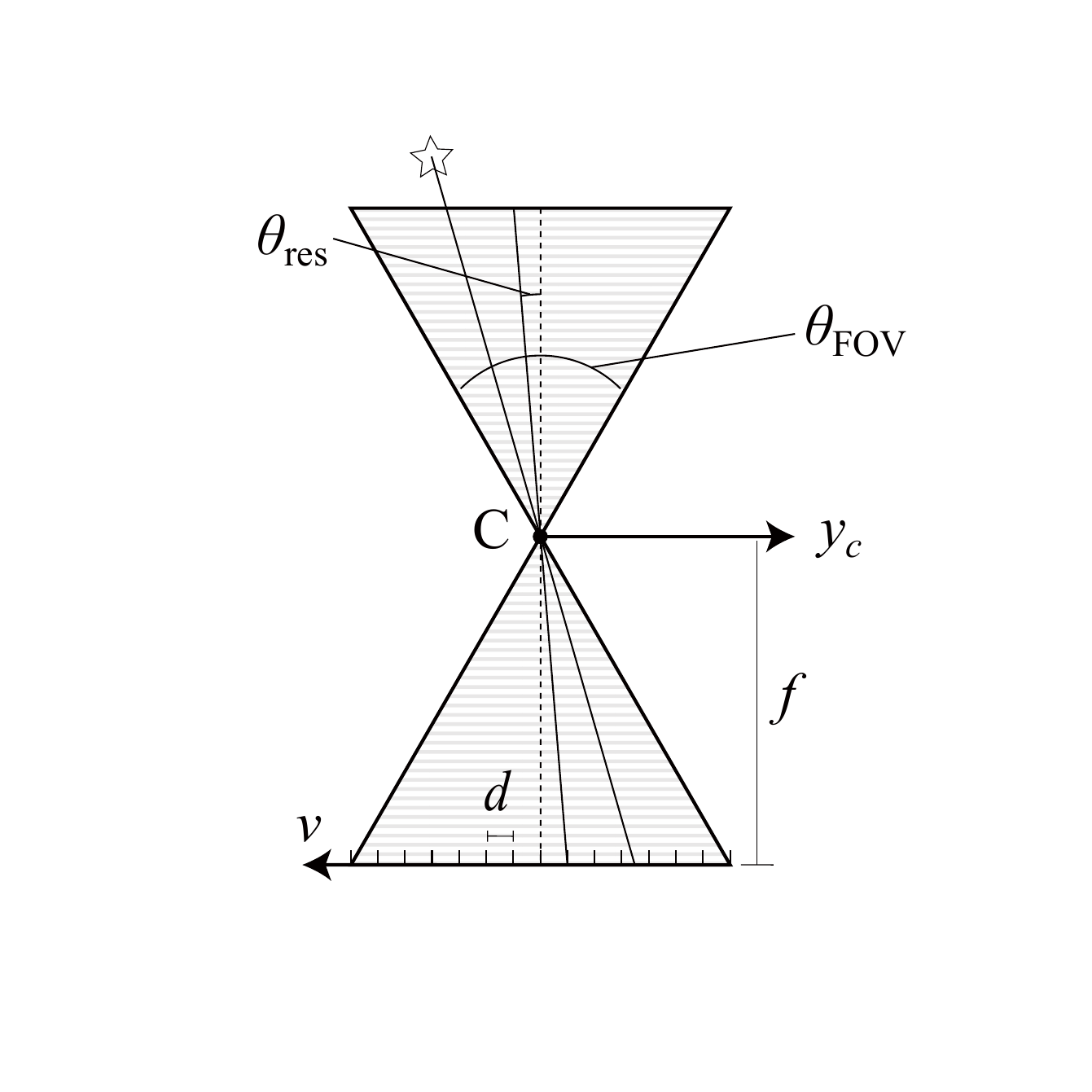}
                \caption{Cross-sectional view of the hatched area in \Cref{fig:camera}}
                \label{fig:camera_cross}
            \end{figure}
    
            As shown in \Cref{fig:camera_cross}, let $\theta_{\mathrm{FOV}} \in \mathbb{S}$ represent the horizontal (vertical) angle of view, and $d$ denote the pixel pitch on the image sensor. The relationship between the pixel pitch $d$ and the angle of view $\theta_{\mathrm{FOV}}$ is given by:
            \begin{equation}
                \frac{Ud}{2} = f \tan{\left( \frac{\theta_{\mathrm{FOV}}}{2} \right)}
                \label{eq:pixlen_fov}
            \end{equation}
            Here, considering that the number of pixels is even, the angular resolution per pixel is largest for pixels that contain the principal point at their vertex. The maximum angular resolution per pixel denoted as $\theta_{\mathrm{res}} \in \mathbb{S}$ satisfies the equation:
            \begin{equation}
                d = f \tan{\left( \theta_{\mathrm{res}} \right)}
                \label{eq:max_pixlen_fov}
            \end{equation}
            By eliminating $d$ from \Cref{eq:pixlen_fov,eq:max_pixlen_fov}, the maximum angular resolution $\theta_{\mathrm{res}}$ is expressed as:
            \begin{equation}
                \theta_{\mathrm{res}} = \tan^{-1}{\left(\frac{2 \tan{\left(\theta_{\mathrm{FOV}}/2\right)}}{U}\right)}
                \label{eq:theta_max}
            \end{equation}
            In this pinhole camera model, the angular resolution at any pixel on the image sensor is always less than or equal to the maximum angular resolution $\theta_{\mathrm{res}}$.
    
\begin{figure}[t]
                \centering
                \includegraphics[width=\hsize]{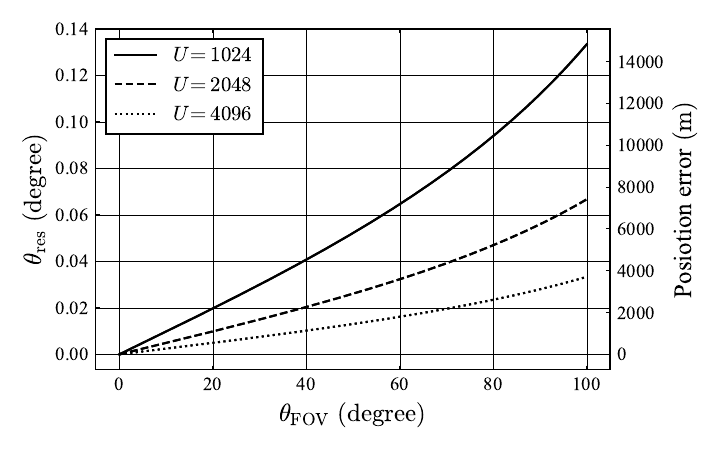}
                \caption{Variation of approximate angular resolution with respect to viewing angle and number of pixels. The right-hand vertical axis shows the distance at sea converted from $\theta_{\mathrm{res}}$.}
                \label{fig:resolutions}
            \end{figure}

            \Cref{fig:resolutions} shows the approximate maximum angular resolution $\theta_{\mathrm{res}}$ corresponding to various FOV angles and the number of pixels, as obtained from \Cref{eq:theta_max}. Additionally, we relate $1/60$ degree to 1 nautical mile, converting the angular distance to sea distance on the right vertical axis.
    
    
            In this study, we provided a simple estimate of the angular resolution as a reference for the angular errors considered in \Cref{sec:starid}. However, it should be noted that in practice, the angular resolution may vary for pixels near the edges due to lens distortion. Additionally, centroiding techniques may improve measurement accuracy.

    \section{Influence of FOV on star identification}\label{sec:starid}    
        In the field of aerospace, many research findings on star ID methods have been published \cite{Spratling2009, Christian2021}. The features used for star ID include magnitude and position of stars. However, due to the variability in image sensor characteristics, the measured brightness of stars may not match the magnitude known in advance. Additionally, at sea, factors such as weather conditions and atmospheric states can also influence the measurements. Therefore, it is challenging to use magnitude, and many star ID methods rely on the relative directional relationships of stars. star ID methods based on the relative positions of stars can be broadly categorized into two types: subgraph isomorphism problem-based methods \cite{Liebe1993, MORTARI2004, Christian2021crossratio, Zhang2017-3} and pattern recognition problem-based methods \cite{Padgett1997, Lee2007, Na2009, Zhang2017-4}. The former is easy to implement but requires a large amount of memory and a long search time, which are considered disadvantages. The latter requires relatively less memory and is compatible with neural networks \cite{BARDWELL1995, Hong2000, ZHANG2008}, which have achieved many successes in the field of pattern recognition.
        
        In this study, we use subgraph isomorphism problem-based methods. One reason for this choice is that, assuming operation on ships, the disadvantages of large memory requirements and long search times of subgraph isomorphism problem-based methods are not critical. Additionally, these methods can easily accommodate an increase in the FOV, making them more practical compared to pattern recognition problem-based methods.
            
        In this section, to discuss the probability of correct matching based on the size of the FOV, we present the results of star ID simulations using subgraph isomorphism-based methods. These simulations take into account the measurement accuracy discussed in \Cref{sec:measure}, as well as the effects of the reduction in observable star magnitude and the presence of stars covered by clouds or the moon.
    
        \subsection{Spherical coordinates}\label{subsec:sph_coo}
        
            \begin{figure}[t]
                \centering
                \includegraphics[width=0.6\hsize]{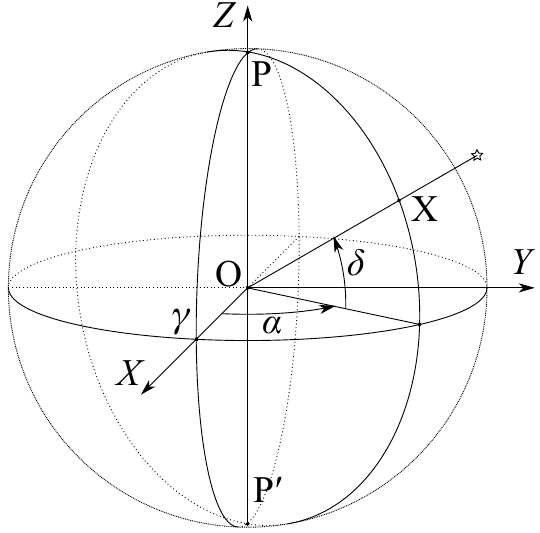}
                \caption{Equatorial coordinate systems.}
                \label{fig:equatorial}
            \end{figure}
        
            Stars can be considered to be located at an infinite distance from the observation point. Therefore, it can be assumed that all stars exist on a virtual sphere centered on the earth. Thus, the position of stars can be represented by unit vectors in either the spherical coordinate system or the orthogonal coordinate system.
        
            One of the spherical coordinate systems used to represent the position of stars is the equatorial coordinate system. The equatorial coordinate system is widely used to specify the positions of celestial bodies. As shown in \Cref{fig:equatorial}, consider a right-handed 3D orthogonal coordinate system $\mathrm{O}$-$XYZ$ with the the centre of Earth as the origin. In the equatorial coordinate system, the origin is the center of the earth $\mathrm{O}$, the fundamental plane is the $XY$-plane, and the principal direction is the positive direction of the $X$ axis. In this system, the positions of celestial bodies are expressed in terms of right ascension and declination, denoted by $\alpha$ and $\delta$, respectively. Right ascension $\alpha$ represents the angle between the vernal equinox $\gamma$ and the hour circle of the star $\mathrm{X}$, and declination $\delta$ represents the angle between the star $\mathrm{X}$ and the celestial equator. The hour circle of the star $\mathrm{X}$ is a celestial circle passing through the star $\mathrm{X}$, the celestial north pole $\mathrm{P}$, and the celestial south pole $\mathrm{P}^{\prime}$. The vernal equinox $\gamma$ is one of the two points where the ecliptic intersects the celestial equator.
    
            Another spherical coordinate system used in this paper is the galactic coordinate system. In the galactic coordinate system, the Sun is the origin, the fundamental plane is parallel to the galactic plane, and the primary direction points toward the approximate center of the Milky Way galaxy. The positions of celestial bodies in this system are expressed in terms of galactic latitude and galactic longitude, denoted by $b$ and $\ell$, respectively. The conversion from equatorial coordinates to galactic coordinates is given by the following equations \cite{Binney1998}:
            \begin{equation}
                \left\{\begin{aligned}
                    \sin b & = \cos \delta \cos \delta_G \cos \left(\alpha-\alpha_G\right)+\sin \delta \sin \delta_G \\
                    \sin \left(\ell_{N}-\ell\right) \cos b & =  \cos \delta \sin \left(\alpha-\alpha_G\right) \\
                    \cos \left(\ell_{N}-\ell\right) \cos b & = \sin \delta \cos \delta_G-\cos \delta \sin \delta_G \cos \left(\alpha-\alpha_G\right)
                \end{aligned}\right.
                \label{eq:equatorialtogalactic}
            \end{equation}
            Here, $\alpha_G$ and $\delta_G$ denote the right ascension and declination of the north galactic pole, respectively, and $\ell_{N}$ represents the galactic longitude of the celestial north pole. The values are set as $\alpha_G=192^{\circ}.85948$, $\delta_G=27^{\circ}.12825$, and $\ell_{N}=122^{\circ}.93192$ \cite{Hipparcos1997}.
            
            In this paper, the position of a star in the $\mathrm{O}$-$XYZ$ coordinate system is represented as:
            \begin{equation}
                \boldsymbol{s} = \begin{pmatrix}
                \cos \alpha \cos \delta \\
                \sin \alpha \cos \delta \\
                \sin \delta
                \end{pmatrix}
                \label{eq:equatorialcoo}
            \end{equation}
            If the positions of two stars are represented by $\boldsymbol{s}_{i} \in \mathbb{R}^{3}$ and $\boldsymbol{s}_{j} \in \mathbb{R}^{3}$, the angular distance between these stars $\Theta\left( \boldsymbol{s}_{i} \cdot \boldsymbol{s}_{j} \right) \in \mathbb{S}$ is given by:
            \begin{equation}
                \Theta\left( \boldsymbol{s}_{i} \cdot \boldsymbol{s}_{j} \right) = \cos^{-1} \left( \frac{\boldsymbol{s}_{i} \cdot \boldsymbol{s}_{j}}{\left|\boldsymbol{s}_{j}\right|\left|\boldsymbol{s}_{j}\right|} \right)
                \label{eq:interangle}
            \end{equation}
            This angular distance can be determined not only from equatorial coordinates but also from pixel coordinates using \Cref{eq:pixel}.
    
        \subsection{Subgraph isomorphism based method}\label{subsec:subgraph}
\begin{figure}[t]
                \centering
                \includegraphics[width=\hsize]{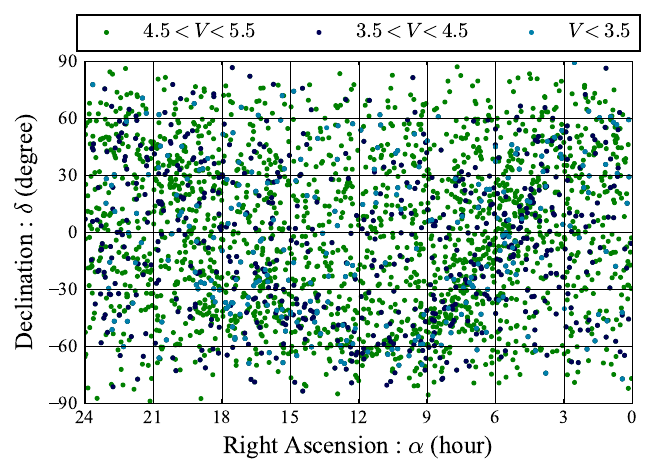}
                \caption{Distribution of stars below magnitude 5.5 in equatorial coordinate systems}
                \label{fig:star_distribution}
            \end{figure}

\begin{figure}[t]
                \centering
                \includegraphics[width=\hsize]{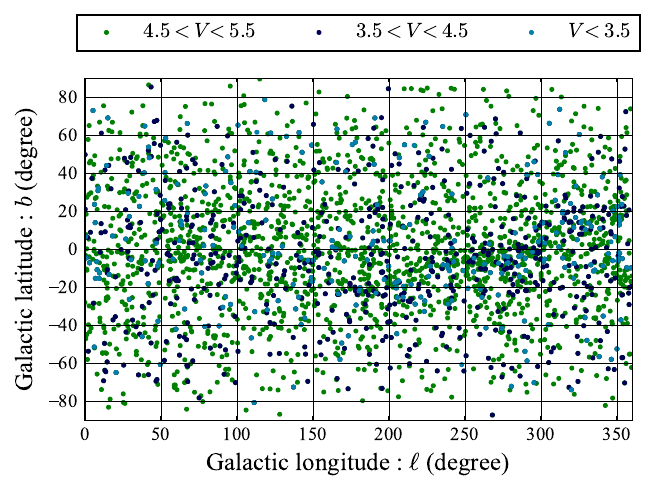}
                \caption{Distribution of stars below magnitude 5.5 in galactic coordinate systems}
                \label{fig:star_distribution_gala}
            \end{figure}

            Star ID is performed by finding stars in a star catalog that match the features of the measured stars. A star catalog is a catalog that lists identification numbers, positions, visual magnitudes, and other information about stars. Here, the visual magnitude is the magnitude of the star in yellow-green light and is the centre of the wavelength range that can be recognised by the eye. Representative star catalogs include the Hipparcos astrometric catalog \cite{HIPPARCOS} and the Yale bright star catalog \cite{yale}. The Yale bright star catalog records stars brighter than magnitude 6.5, and this paper uses this catalog. The distribution of stars brighter than magnitude 5.5 in equatorial coordinates is shown in \Cref{fig:star_distribution}, and in galactic coordinates in \Cref{fig:star_distribution_gala}.
    
        
            In the subgraph isomorphism-based method, stars are treated as vertices, and the angular distances between stars are treated as edges, thus considering the set of stars as a graph. The graph of the observed set of stars is considered a subgraph of the graph formed by the set of all observable stars. In other words, the identification of the measured stars is performed by graph matching, finding the subgraph that matches the observed graph.
    
            \subsubsection{Database}
                Most subgraph isomorphism-based methods prepare an angular distance database in advance, which records the inter-star angle values for combinations of observable stars, in addition to the star catalog, to efficiently compute graph matching. In this study, we prepare a modified star database and a database recording the angular distances between pairs of stars.
                    
                Let the number of stars recorded in the star catalog be $N_{\mathrm{C}}$. The identification number, position, and visual magnitude of the $i$-th star are denoted as $I_{i}, \boldsymbol{s}_{i}, V_{i}$, respectively. Note that $\boldsymbol{s}_{i}$ is a unit vector calculated from the right ascension and declination using \Cref{eq:equatorialcoo}. The set derived from the star catalog is given by:
                \begin{equation}
                    \mathcal{D}_{\mathrm{C}} = \left\{ (I_{i}, \boldsymbol{s}_{i}, V_{i}) \mid i \in [1, 2, \cdots, N_{\mathrm{C}}] \right\}
                    \label{eq:dataset_catalog}
                \end{equation}
                For simplicity, we denote $(I, \boldsymbol{s}, V)$ as $D$ in this paper.
                To avoid misidentification, stars that are too faint to be observed by the equipment are excluded from the set. Letting the upper limit of visual magnitude be $M_{\mathrm{lim}}$, the set after exclusion is given by:
                \begin{equation}
                    \mathcal{D}^{\prime} = \left\{ D_{\in \mathcal{D}_{\mathrm{C}}} \mid V \leq M_{\mathrm{lim}} \right\}
                    \label{eq:dataset_Mlim}
                \end{equation}
                Furthermore, binary stars whose inter-star distance is closer than the instrumental resolution are also excluded. Let $\theta_{\mathrm{min}} \in \mathbb{S}$ denote the threshold of angular distance. The set of stars in $\mathcal{D}^{\prime}$ whose angular distance from any other celestial body in $\mathcal{D}^{\prime}$ is greater than $\theta_{\mathrm{min}}$ is given by:
                \begin{equation}
                    \mathcal{D}_{\mathrm{DB}} = \left\{ D_{\in \mathcal{D}^{\prime}} \mid \forall {D^{\prime}}_{\in \mathcal{D}^{\prime}}\left[\left(\theta_{\mathrm{min}} \leq \Theta\left(\boldsymbol{s}, \boldsymbol{s}^{\prime}\right)\right) \lor \left(I = I^{\prime}\right) \right] \right\}
                    \label{eq:dataset}
                \end{equation}
                Thus, $\mathcal{D}_{\mathrm{DB}}$ becomes the modified star database.
        
                When using an observation device with a fixed FOV, the maximum value that the inter-star angle can take is fixed. Let $\theta_{\mathrm{max}} \in \mathbb{S}$ denote the maximum inter-star angle. All pairs of stars with inter-star angles smaller than $\theta_{\mathrm{max}}$ are represented as:
                \begin{equation}
                    \begin{aligned} 
                        \mathcal{P}_{\mathrm{DB}} = \Bigl\{ \Bigr.&\left(D, D^{\prime}, \Theta\left(\boldsymbol{s}, \boldsymbol{s}^{\prime}\right) \right) \mid  \\
                        & \Bigl. (D, D^{\prime}) \in \mathcal{D}_{\mathrm{star}}^2 \land \left(\Theta\left(\boldsymbol{s}, \boldsymbol{s}^{\prime}\right) \leq \theta_{\mathrm{max}}\right) \land \left(I \neq I^{\prime}\right) \Bigr\}
                    \end{aligned}
                    \label{eq:pairset}
                \end{equation}
                Therefore, $\mathcal{P}_{\mathrm{DB}}$ is used as the angular distance database.
                
            \subsubsection{Matching algorithm}
                In graph matching, we use the prepared database to search for subgraphs that match the observed graph. The directional relationships used in matching include the angular distances and internal angles of three stars \cite{Liebe1993, Zhang2017-3}, the angular distances of a pyramid formed by four stars \cite{MORTARI2004}, and the cross-ratio of five stars \cite{Christian2021crossratio}. The Pyramid algorithm using four stars \cite{MORTARI2004} is a representative matching method known for its high identification accuracy. 
                In this study, we simulate matching using each angular distance of an arbitrary number of stars. The matching procedure is explained below.
                
                First, consider the situation where two measured stars are given. The positions of each star are denoted as $\hat{\boldsymbol{s}}_{1} \in \mathbb{R}^{3}$ and $\hat{\boldsymbol{s}}_{2} \in \mathbb{R}^{3}$. Note that $\hat{\boldsymbol{s}}_{1}$ and $\hat{\boldsymbol{s}}_{2}$ are unit vectors in the coordinate system $\mathrm{C}$-$x_{\mathrm{c}}y_{\mathrm{c}}z_{\mathrm{c}}$ based on the measuring equipment. 
                In this case, pairs of catalog stars whose inter-star angle $\theta \in [0, \theta_{\mathrm{max}}]$ satisfy the following condition are searched from the angular distance database $\mathcal{P}_{\mathrm{DB}}$.
                \begin{equation}
                    \left| \theta - \cos^{-1} \left( \hat{\boldsymbol{s}}_{1} \cdot \hat{\boldsymbol{s}}_{2} \right) \right| \leq \varepsilon
                    \label{eq:candipair}
                \end{equation}
                Here, $\varepsilon$ represents the allowable error for the angular distance. 
                This algorithm is summarized in \Cref{alg:search_two_stars}. The set $\mathcal{C}_{12}$ represents the pairs matching $\hat{\boldsymbol{s}}_{1}$ and $\hat{\boldsymbol{s}}_{2}$.
    
                \begin{algorithm}
                    \caption{Matching procedure of pair stars.}
                    \begin{algorithmic}[1]
                        \Function {match\_$2$\_stars}{$\hat{\boldsymbol{s}}_{1},\hat{\boldsymbol{s}}_{2}, \mathcal{P}_{\mathrm{DB}}, \varepsilon$}
                            \State $\hat{\theta} \gets \cos^{-1} \left( \hat{\boldsymbol{s}}_{1} \cdot \hat{\boldsymbol{s}}_{2} \right)$
                            \State $\mathcal{C}_{12} \gets \left\{  \left(D, D^{\prime}\right) \mid \left(D, D^{\prime}, \theta \right) \in \mathcal{P}_{\mathrm{DB}} \land \left(| \theta - \hat{\theta} | \leq \varepsilon\right) \right\}$
                            \State return $\mathcal{C}_{12}$
                        \EndFunction
                    \end{algorithmic}
                    \label{alg:search_two_stars}
                \end{algorithm}
    
                To identify the measured stars, the number of elements in the matched pair set must be 1, i.e., $\left|\mathcal{C}_{12}\right|=1$. However, it is practically difficult to reduce the allowable error $\varepsilon$ to such an extent that \Cref{alg:search_two_stars} results in only one element. Therefore, by matching more measured stars simultaneously, it is possible to narrow down the candidates.
    
                For example, in the situation where three measured stars are given, matching can be performed by searching for combinations of three stars whose three inter-star angles match simultaneously. The matching algorithm used in this study is defined in \Cref{alg:search_three_stars}. The set $\mathcal{C}_{123}$ represents the set of star combinations matching the graph of the three measured stars $\hat{\boldsymbol{s}}_{1},\hat{\boldsymbol{s}}_{2},\hat{\boldsymbol{s}}_{3}$. In this algorithm, as proposed in the Pyramid algorithm \cite{MORTARI2004}, enantiomers are excluded in line 11 of \Cref{alg:search_three_stars}.
    
                \begin{algorithm}
                    \caption{Matching procedure of triangle stars.}
                    \begin{algorithmic}[1]
                        \Function {match\_$3$\_stars}{$\hat{\boldsymbol{s}}_{1},\hat{\boldsymbol{s}}_{2},\hat{\boldsymbol{s}}_{3}, \mathcal{P}_{\mathrm{DB}}, \varepsilon$}
                            \State $\mathcal{C}_{12} \gets $ \Call{match\_$2$\_stars}{$\hat{\boldsymbol{s}}_{1},\hat{\boldsymbol{s}}_{2}, \mathcal{P}_{\mathrm{DB}}, \varepsilon$}
                            \State $\mathcal{C}_{23} \gets $ \Call{match\_$2$\_stars}{$\hat{\boldsymbol{s}}_{2},\hat{\boldsymbol{s}}_{3}, \mathcal{P}_{\mathrm{DB}}, \varepsilon$}
                            \State $\mathcal{C}_{13} \gets $ \Call{match\_$2$\_stars}{$\hat{\boldsymbol{s}}_{1},\hat{\boldsymbol{s}}_{3}, \mathcal{P}_{\mathrm{DB}}, \varepsilon$}
                            \State $\mathcal{D}_{1}^{\mathrm{(candi)}} \gets \bigcup_{\left(D, D^{\prime} \right) \in C_{12} \cup C_{13}}^{} \{D, D^{\prime}\}$
                            \State $\mathcal{C}_{123}^{\mathrm{(candi)}} \gets \emptyset$
                            \ForAll {$D \in \mathcal{D}_{1}^{\mathrm{(candi)}}$}
                                \State $\mathcal{D}_{2}^{\mathrm{(candi)}} \gets \left\{ D^{\prime\prime} \mid \left(D^{\prime}, D^{\prime\prime}\right) \in \mathcal{C}_{12} \land (D=D^{\prime}) \right\} \cup $
                                \Statex \hfill $ \left\{ D^{\prime} \mid \left(D^{\prime}, D^{\prime\prime} \right) \in \mathcal{C}_{12} \land (D=D^{\prime\prime}) \right\}$
                                \State $\mathcal{D}_{3}^{\mathrm{(candi)}} \gets \left\{ D^{\prime\prime} \mid \left(D^{\prime}, D^{\prime\prime}\right) \in \mathcal{C}_{13} \land (D=D^{\prime}) \right\} \cup $
                                \Statex \hfill $ \left\{ D^{\prime} \mid \left(D^{\prime}, D^{\prime\prime} \right) \in \mathcal{C}_{13} \land (D=D^{\prime\prime}) \right\}$
                                \State $\mathcal{C}_{123}^{\mathrm{(candi)}} \gets \mathcal{C}_{123}^{\mathrm{(candi)}} \cup \Bigl\{ (D, D^{\prime}, D^{\prime\prime}) \mid \Bigr.$
                                \Statex \hfill $ \left. D^{\prime} \in \mathcal{D}_{2}^{\mathrm{(candi)}} \land D^{\prime\prime} \in \mathcal{D}_{3}^{\mathrm{(candi)}} \land \left(D^{\prime}, D^{\prime\prime}\right) \in \mathcal{C}_{23} \right\}$
                            \EndFor
                            \State $C_{123} \gets \Bigl\{ (D, D^{\prime}, D^{\prime\prime}) \in \mathcal{C}_{123}^{\mathrm{(candi)}} \mid \Bigr.$
                            \Statex \hfill $\Bigl. \mathrm{sign}\left(\boldsymbol{s} \cdot \left(\boldsymbol{s}^{\prime} \times \boldsymbol{s}^{\prime\prime}\right)\right) = \mathrm{sign}\left(\hat{\boldsymbol{s}}_{1} \cdot \left(\hat{\boldsymbol{s}}_{2} \times \hat{\boldsymbol{s}}_{3}\right)\right) \Bigr\}$
                            \State return $C_{123}$
                        \EndFunction
                    \end{algorithmic}
                    \label{alg:search_three_stars}
                \end{algorithm}
    
                Additionally, the number of elements in $\mathcal{C}_{123}$ is not necessarily 1. By matching more measured stars, such as the four used in the Pyramid algorithm \cite{MORTARI2004}, more conditions related to inter-star angles can be added. In the matching problem for $p>3$ measured stars, given a matching set of $p-1$ measured stars, conditions on $p-1$ inter-star angles is added. The matching algorithm used in this study is defined in \Cref{alg:search_p_stars}. The set $\mathcal{C}_{12 \cdots p}$ represents the set of star combinations matching the graph of $p$ measured stars $\hat{\boldsymbol{s}}_{1},\hat{\boldsymbol{s}}_{2}, \cdots, \hat{\boldsymbol{s}}_{p}$.
                        
                \begin{algorithm}
                    \caption{Matching procedure for more stars than 3.}
                    \begin{algorithmic}[1]
                    \Function {match\_$p$\_stars}{$\hat{\boldsymbol{s}}_{1},\hat{\boldsymbol{s}}_{2}, \cdots, \hat{\boldsymbol{s}}_{p}, \mathcal{P}_{\mathrm{DB}}, \varepsilon$}
                        \State $\mathcal{C}_{12 \cdots (p-1)} \gets $
                        \Statex \hfill \Call{match\_$(p-1)$\_stars}{$\hat{\boldsymbol{s}}_{1}$, $\hat{\boldsymbol{s}}_{2}$, $\cdots$, $\hat{\boldsymbol{s}}_{(p-1)}$, $\mathcal{P}_{\mathrm{DB}}, \varepsilon$}
                        \ForAll {$i \in \left( 1, 2, \cdots, p-1\right)$}
                            \State $\mathcal{C}_{ip} \gets $ \Call{match\_$2$\_stars}{$\hat{\boldsymbol{s}}_{i},\hat{\boldsymbol{s}}_{p}, \mathcal{P}_{\mathrm{DB}}, \varepsilon$}
                        \EndFor
                        \State $\mathcal{C}_{12 \cdots p} \gets \emptyset$
                        \ForAll {$\left(D_1, D_2, \cdots, D_{p-1}\right) \in C_{12 \cdots (p-1)}$} 
                            \ForAll {$i \in \left( 1, 2, \cdots, p-1\right)$}
                                \State $\mathcal{D}_{p,i}^{\mathrm{(candi)}} \gets \left\{ D^{\prime} \mid \left(D, D^{\prime}\right) \in \mathcal{C}_{ip} \land (D_{i}=D) \right\} \cup $
                                \Statex \hfill $ \left\{ D \mid \left(D, D^{\prime} \right) \in \mathcal{C}_{ip} \land (D_{i}=D^{\prime}) \right\}$
                            \EndFor
                            \State $\mathcal{D}_{p}^{\mathrm{(candi)}} \gets \bigcap_{i \in \left( 1, 2, \cdots, p-1\right)} \mathcal{D}_{p,i}^{\mathrm{(candi)}}$
                            \State $\mathcal{C}_{12 \cdots p} \gets \mathcal{C}_{12 \cdots p} \cup \left\{ \left(D_1, D_2, \cdots, D_{p}\right) \mid D_{p} \in \mathcal{D}_{p}^{\mathrm{(candi)}} \right\} $
                        \EndFor
                        \State return $\mathcal{C}_{12 \cdots p}$
                    \EndFunction
                    \label{alg:search_p_stars}
                    \end{algorithmic}
                \end{algorithm}
    
                In general, the more inter-star angles that are matched, the fewer elements the set of matched combinations will have. Therefore, even if the observational error of the angular distance is large and $\varepsilon$ cannot be made small, stars can be identified by increasing the number of measured stars to be matched, $p$. However, increasing the number of measured stars $p$ also increases the computation time required for matching.
    
                In this study, we use an identification algorithm that combines \Cref{alg:search_two_stars,alg:search_three_stars,alg:search_p_stars}. Specifically, as shown in \Cref{alg:search_stars}, if the number of measured stars is less than two, the algorithm returns an empty set. Otherwise, it increases the number of measured stars used for matching until the number of elements in the matched combination set is less than or equal to $1$.
    
                \begin{algorithm}
                    \caption{Matching procedure of $p$ stars in this study.}
                    \begin{algorithmic}[1]
                    \Function {match\_stars}{$\hat{\boldsymbol{s}}_{1},\hat{\boldsymbol{s}}_{2}, \cdots, \hat{\boldsymbol{s}}_{p}, \mathcal{P}_{\mathrm{DB}}, \varepsilon$}
                        \If {$p \leq 1$}
                            \State return $\emptyset$
                        \EndIf
                        \ForAll {$i \in \left( 2, \cdots, p\right)$}
                            \State $\mathcal{C}_{12 \cdots i} \gets $ \Call{match\_$i$\_stars}{$\hat{\boldsymbol{s}}_{1},\hat{\boldsymbol{s}}_{2},\cdots,\hat{\boldsymbol{s}}_{i}, \mathcal{P}_{\mathrm{DB}}, \varepsilon$}
                            \If {$\left|\mathcal{C}_{12 \cdots i}\right|=1$}
                                \State return $\mathcal{C}_{12 \cdots i}$
                            \EndIf
                        \EndFor
                        \State return $\mathcal{C}_{12 \cdots p}$
                    \EndFunction
                    \label{alg:search_stars}
                    \end{algorithmic}
                \end{algorithm}
    
                If the number of measured stars is small, the number of elements in the matched combination set $\mathcal{C}_{12 \cdots p}$ from \Cref{alg:search_stars} may not be determined as one. In this case, even if the correct match is included in the candidate matching set, measured stars cannot be identified. Therefore, a ``correct matching'' is defined as the event where the number of elements in the matched combination set $\mathcal{C}_{12 \cdots p}$ is one and the correct match is included in the matched combination set.
                
        \subsection{Identification simulation including the ship's environment}\label{subsec:sim}
            Here, we present the number of observable stars and the probability of correct matching considering the ship's environment using Monte Carlo simulations. The following influences are considered in this simulation:
            \begin{itemize}
                \item \textbf{Measurement accuracy based on FOV size}: The measurement accuracy determined based on the angular resolution with a pixel count of $U=1,024$, as obtained in \Cref{sec:measure}, is considered.
                \item \textbf{Decrease in observable magnitude}: The effects of reduced exposure time due to ship motion and decreased magnitude of stars due to atmospheric conditions are considered. In the simulation, this is represented by reducing the maximum observable magnitude.
                \item \textbf{Presence of covered stars}: The possibility of stars being obscured by clouds or the moon is considered. In the simulation, this is represented by randomly removing measured stars at a certain rate.
            \end{itemize}
            The second and third items are introduced to account for the impacts caused by environmental disturbances in the ship's environment. The purpose of this simulation is to clarify the probability of correct matching concerning the FOV size, considering the effects of the ship's environment.
    
            \subsubsection{Simulation procedure}\label{subsubsec:procedure}
                We describe the procedure for the Monte Carlo simulations conducted in this study. In this simulation, the attitude of the measuring equipment is randomly determined, and the observable stars are determined by whether the star is in the FOV or not. It is assumed that all stars with a magnitude of $\hat{M}_{\mathrm{lim}}$ or less, included in the Star Database $\mathcal{D}_{\mathrm{DB}}$ and within the FOV, are observed. Star ID is then performed on the measured stars using \Cref{alg:search_stars}. The specific procedure is as follows:
                \begin{enumerate}
                    \item Randomly select a rotation matrix $R \in SO(3)$.
                    \item Transform the position coordinates $\boldsymbol{s}$ of all stars in $\mathcal{D}_{\mathrm{DB}}$ from the $\mathrm{O}$-$XYZ$ coordinate system to the position coordinates $\boldsymbol{s}_{\mathrm{c}}=R\boldsymbol{s}$ in the $\mathrm{C}$-$x_{\mathrm{c}}y_{\mathrm{c}}z_{\mathrm{c}}$ coordinate system.
                    \item Select all stars as measured stars if their transformed positions $\boldsymbol{s}_{c}$ and magnitudes $V$ satisfy the following conditions:
                        \begin{equation}
                            \left\{\begin{aligned}
                                s_{\mathrm{c}1}/s_{\mathrm{c}3} &< \tan\left(\theta_{\mathrm{FOV}}/2\right) \\
                                s_{\mathrm{c}2}/s_{\mathrm{c}3} &< \tan\left(\theta_{\mathrm{FOV}}/2\right) \\
                                s_{\mathrm{c}3} &> 0.0 \\
                                V &< \hat{M}_{\mathrm{lim}} \\
                            \end{aligned}\right.
                            \label{eq:obs_condi}
                        \end{equation}
                        where $(s_{\mathrm{c}1}, s_{\mathrm{c}2}, s_{\mathrm{c}3})$ are the components of $\boldsymbol{s}_{\mathrm{c}}$.
                    \item Add random angular errors of up to $\varepsilon/2$ as observation errors to the positions of the measured stars.
                    \item Exclude measured stars with a probability of $\beta$.
                    \item Randomly reorder the measured stars.
                    \item Obtain the set of matching candidates using \Cref{alg:search_stars}.
                \end{enumerate}
                In this simulation, for each trial, the following were recorded: the galactic latitude $b_{c}$ of the measuring equipment's orientation (i.e., the $z_{c}$ axis direction), the number of observable stars $N$, whether the matching was correct, and the number of stars $p_{\mathrm{match}}$ required for matching.

                The number of trials in the Monte Carlo simulation was set to $20,000$. The Monte Carlo simulation was conducted for all combinations of the parameters representing ship environment disturbances $(\theta_{\mathrm{FOV}}, \hat{M}_{\mathrm{lim}}, \beta) \in \{5^{\circ}, 10^{\circ}, 20^{\circ}, 40^{\circ}, 80^{\circ}\} \times \{3.5, 4.5, 5.5\} \times \{0.0, 0.2, 0.4, 0.6, 0.8\}$. Here, $\theta_{\mathrm{res}}$ was determined using \Cref{eq:theta_max} with $U=1,024$. The Star Database $\mathcal{D}_{\mathrm{DB}}$ and the angular distance database $\mathcal{P}_{\mathrm{DB}}$ were computed with $\theta_{\mathrm{min}} = 2\sqrt{2} \times \theta_{\mathrm{res}}$ and $\theta_{\mathrm{max}}=2 \tan^{-1}\left(\sqrt{2}\tan\left(\frac{\theta_{\mathrm{FOV}}}{2}\right)\right)$. In \Cref{alg:search_stars}, $\varepsilon = 2\sqrt{2} \times \theta_{\mathrm{res}}$ was used. The values of $\varepsilon$ and $\theta_{\mathrm{min}}$ were determined based on the consideration that the observation error of a celestial body could be up to $\sqrt{2}$ times the resolution. The value of $\theta_{\mathrm{max}}$ represents the maximum angle that can be taken within the FOV.
        
                Furthermore, to clarify the differences in results due to the density of stars, the simulation was divided based on whether the camera's direction was oriented around the Milky Way, where the density of stars is relatively high. Specifically, if the absolute value of the galactic latitude $b_{c}$ of the camera's direction was less than $30^{\circ}$, the camera was considered to be oriented around the Milky Way. Otherwise, it was considered not to be oriented around the Milky Way.
                
            \subsubsection{Simulation results}\label{sec:sim}
\begin{figure*}[t]
                    \begin{minipage}[c]{\linewidth}
                        \centering
                        \includegraphics[width=\hsize]{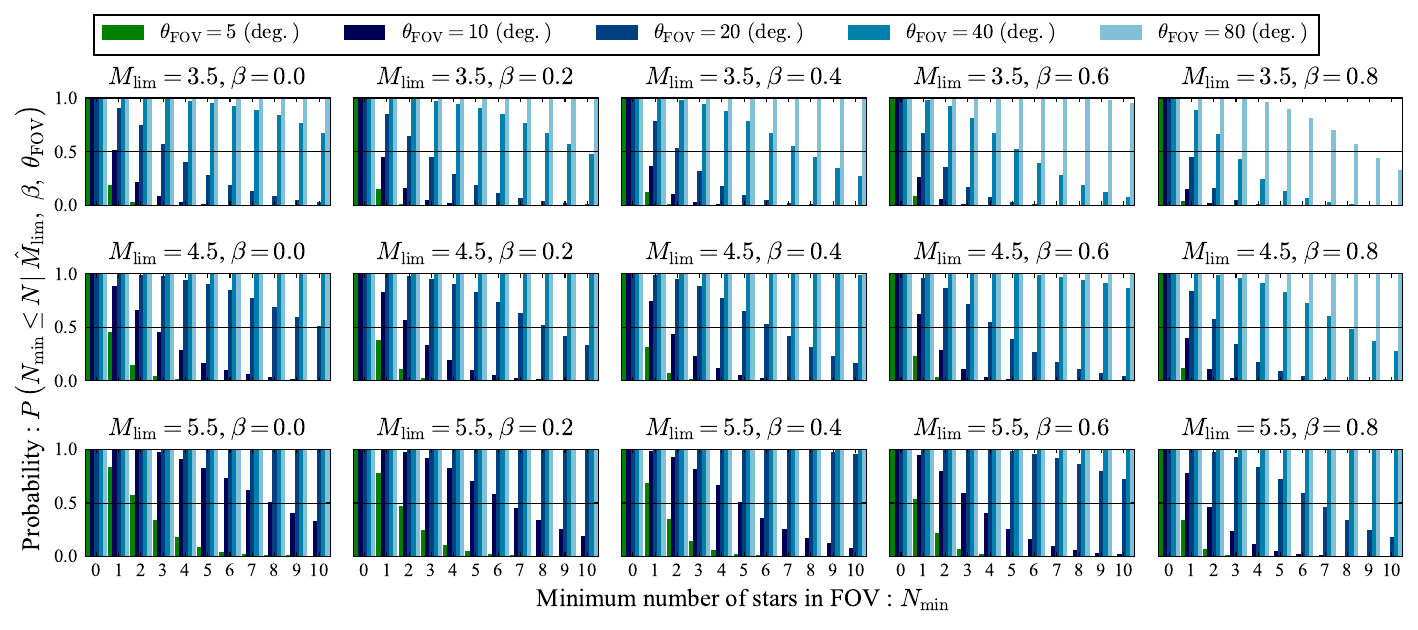}
                        \subcaption{Probability when the camera direction is close to the Milky Way $\left(\left|b_{c}\right| \leq 30^{\circ}\right)$.}
                        \label{fig:obs_cum_distribution_bar_milkyTrue}
                    \end{minipage}
                    \begin{minipage}[c]{\linewidth}
                        \centering
                        \includegraphics[width=\hsize]{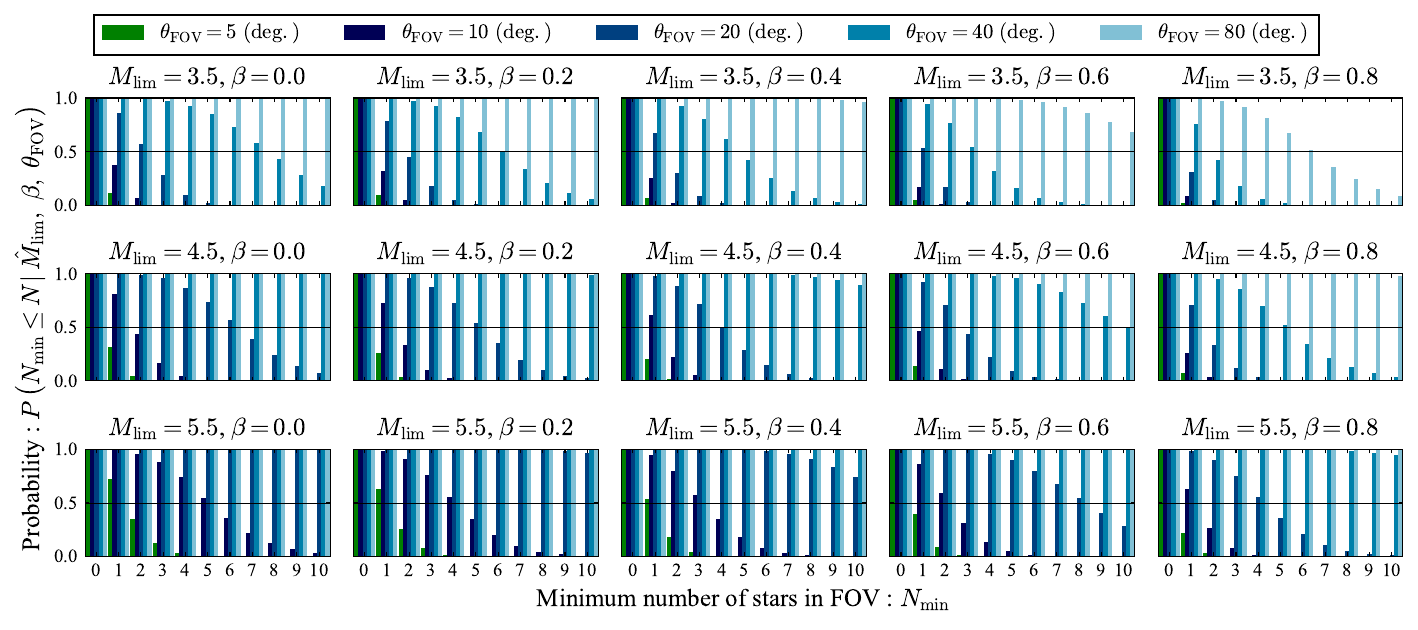}
                        \subcaption{Probability when the camera direction is not close to the Milky Way $\left(\left|b_{c}\right| > 30^{\circ}\right)$.}
                        \label{fig:obs_cum_distribution_bar_milkyFalse}
                    \end{minipage}
                    \caption{Probability of observing at least $N_{\mathrm{min}}$ stars with respect to $\theta_{\mathrm{FOV}}, \hat{M}_{\mathrm{lim}}, \beta$.}
                    \label{fig:obs_cum_distribution_bar}
                \end{figure*}

                First, the probability $P\left(N_{\mathrm{min}} \leq N \mid \hat{M}_{\mathrm{lim}}, \theta_{\mathrm{FOV}}, \beta\right)$ of observing more than $N_{\mathrm{min}}$ stars obtained from the simulation is shown in \Cref{fig:obs_cum_distribution_bar}.
                In the Pyramid algorithm \cite{MORTARI2004}, observing four or more stars is required. From \Cref{fig:obs_cum_distribution_bar}, we can see that with $\theta_{\mathrm{FOV}}=80^{\circ}$, except for the case of $\hat{M}_{\mathrm{lim}}=3.5, \beta=0.8$, the probability of not observing four or more stars is low, regardless of whether the camera's direction is around the Milky Way. With $\theta{\mathrm{FOV}}=40^{\circ}$, except for the cases of $\hat{M}_{\mathrm{lim}}=4.5, \beta=0.8$ or $\hat{M}_{\mathrm{lim}}=3.5, \beta \geq 0.2$, and with $\theta_{\mathrm{FOV}}=20^{\circ}$, except for the cases of $\hat{M}_{\mathrm{lim}}=5.5, \beta=0.8$ or $\hat{M}_{\mathrm{lim}} \leq 4.5$, the probability of not observing four or more stars is low even if the camera's direction is not towards the Milky Way. With $\theta_{\mathrm{FOV}}=5^{\circ}, 10^{\circ}$, the probability of not observing four or more stars is high in many cases. Therefore, the probability of observing four or more stars increases as $\theta_{\mathrm{FOV}}$ increases, and with $\theta_{\mathrm{FOV}}=80^{\circ}$, it is possible to observe in most cases.
        
\begin{figure*}[t]
                    \begin{minipage}[c]{0.49\linewidth}
                        \centering
                        \includegraphics[width=\columnwidth]{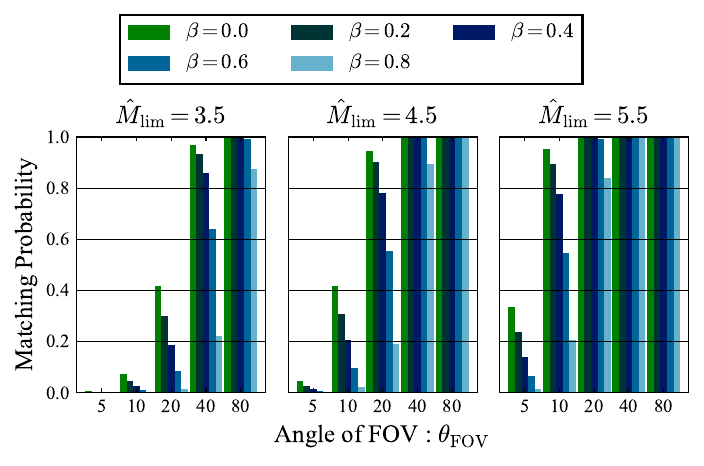}
                        \subcaption{Probabilities of correct matching when the camera direction is close to the Milky Way $\left(\left|b_{c}\right| \leq 30^{\circ}\right)$.}
                        \label{fig:matching_prob_bar_milkyTrue}
                    \end{minipage}
                    \hspace{0.02\columnwidth} 
                    \begin{minipage}[c]{0.49\linewidth}
                        \centering
                        \includegraphics[width=\columnwidth]{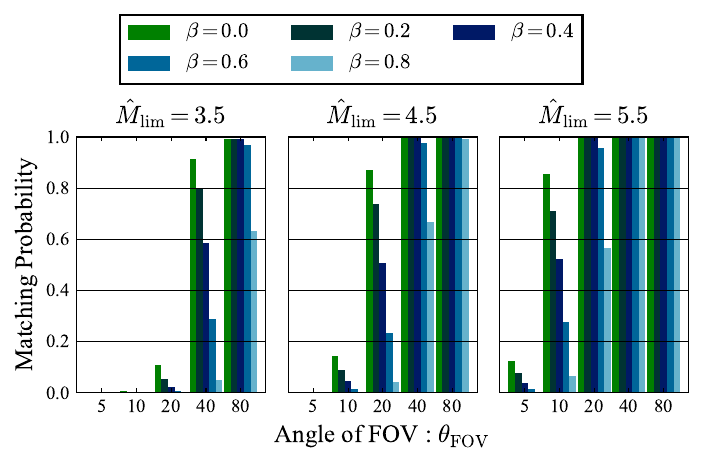}
                        \subcaption{Probabilities of correct matching when the camera direction is not close to the Milky Way $\left(\left|b_{c}\right| > 30^{\circ}\right)$.}
                        \label{fig:matching_prob_bar_milkyFalse}
                    \end{minipage}
                    \caption{Probabilities of correct matching by \Cref{alg:search_stars} with respect to $\theta_{\mathrm{FOV}}, \hat{M}_{\mathrm{lim}}, \beta$.}
                    \label{fig:matching_prob_bar}
                \end{figure*}

                Next, \Cref{fig:matching_prob_bar} shows the probability of correct matching. From \Cref{fig:matching_prob_bar}, it can be seen that with $\theta_{\mathrm{FOV}}=80^{\circ}$, except for the case of $\hat{M}_{\mathrm{lim}}=3.5, \beta=0.8$, the probability of correct matching is high regardless of whether the camera's direction is around the Milky Way. With $\theta{\mathrm{FOV}}=40^{\circ}$, except for the cases of $\hat{M}_{\mathrm{lim}}=4.5, \beta=0.8$ or $\hat{M}_{\mathrm{lim}}=3.5, \beta \geq 0.2$, and with $\theta_{\mathrm{FOV}}=20^{\circ}$, except for the cases of $\hat{M}_{\mathrm{lim}}=5.5, \beta=0.8$ or $\hat{M}_{\mathrm{lim}} \leq 4.5$, the probability of correct matching is high even if the camera's direction is not towards the Milky Way. With $\theta_{\mathrm{FOV}}=5^{\circ}, 10^{\circ}$, the probability of correct matching is low in most cases.
    
\begin{figure*}[t]
                    \centering
                    \includegraphics[width=\hsize]{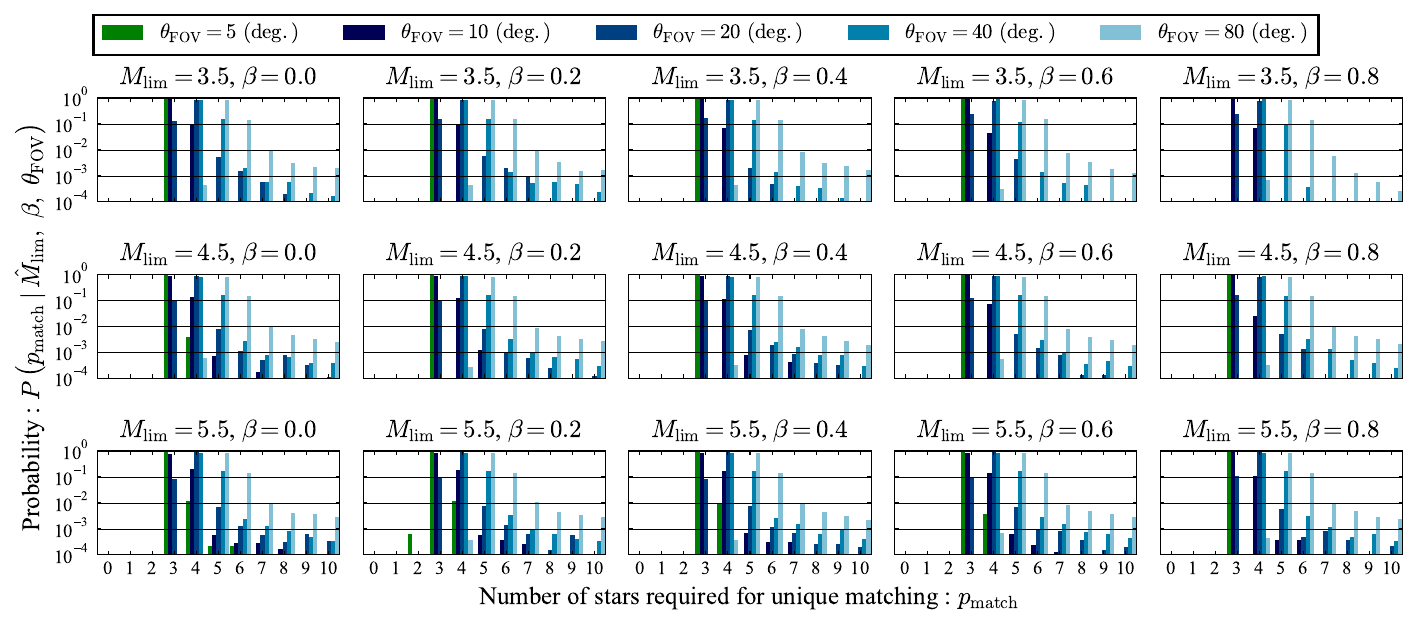}
                    \caption{Probability mass function of number of stars required for unique matching with respect to $\theta_{\mathrm{FOV}}, \hat{M}_{\mathrm{lim}}, \beta$.}
                    \label{fig:required_distribution}
                \end{figure*}

                Finally, \Cref{fig:required_distribution} shows the probability mass distribution $P\left(p_{\mathrm{match}} \mid \hat{M}_{\mathrm{lim}}, \theta_{\mathrm{FOV}}, \beta\right)$ of the number of measured stars required for correct matching $p_{\mathrm{match}}$. Note that the probability mass distribution is not shown for cases with zero probabilities of correct matching, e.g., $\theta_{\mathrm{FOV}}=5^{\circ}, \hat{M}_{\mathrm{lim}}=3.5, \beta=0.8$. From \Cref{fig:required_distribution}, it can be seen that with $\theta_{\mathrm{FOV}}=80^{\circ}$, $p_{\mathrm{match}}=5$; with $\theta_{\mathrm{FOV}}=20^{\circ}, 40^{\circ}$, $p_{\mathrm{match}}=4$; and with $\theta_{\mathrm{FOV}}=5^{\circ}, 10^{\circ}$, $p_{\mathrm{match}}=3$ are the most common. In particular, with $\theta_{\mathrm{FOV}}=80^{\circ}$, the proportion of $p_{\mathrm{match}}=3$ is zero, and the proportion of $p_{\mathrm{match}}=4$ is low, indicating that five or more measured stars are required for high-precision identification.
        
    \section{Discussion}\label{sec:discuss}
        In this section, we discuss the performance specifications required for an automatic celestial navigation system in a maritime environment based on the simulation results obtained in \Cref{sec:measure,sec:starid} from the perspective of FOV size. The changes found from the simulation results with respect to FOV size are summarized as follows:
        \begin{itemize}
            \item From the results in \Cref{fig:resolutions}, it was found that as the FOV increases, the angular resolution decreases, resulting in lower measurement accuracy.
            \item From the results in \Cref{fig:obs_cum_distribution_bar}, it was found that as the FOV increases, the number of observable stars increases.
            \item From the results in \Cref{fig:matching_prob_bar}, it was found that as the FOV increases, the matching probability increases.
            \item From the results in \Cref{fig:required_distribution}, it was found that as the FOV increases, the number of stars required for matching increases.
        \end{itemize}
        From these results, it can be inferred that the increase in the number of observable stars due to a larger FOV has a more significant impact on the star ID probability than the decrease in measurement accuracy. In any case, it was found that increasing the FOV can improve the probability of correct matching of star ID using subgraph isomorphism-based methods. This finding is valid even in scenarios that consider a maritime environment, where the number of observable stars decreases due to atmospheric conditions, clouds, and ship motion.
        
        Moreover, from the probability of observing four or more stars shown in \Cref{fig:obs_cum_distribution_bar} and the probability of correct matching shown in \Cref{fig:matching_prob_bar}, it is found that observing four or more stars is required for accurate identification. However, from the results in \Cref{fig:required_distribution}, it was found that when the FOV is increased, four stars may be insufficient, at least for $\theta_{\mathrm{FOV}} \geq 80^{\circ}$. Therefore, when increasing the FOV, it is necessary to introduce a new algorithm that incorporates methods improving matching efficiency \cite{Mortari1997}, a pattern shift algorithm to avoid false stars \cite{Mueller2016}, and an analytical method for the frequency of star pattern mismatching \cite{Kumar2010} into our algorithm that uses more stars.
        
        As mentioned above, in the shipboard environment, there are almost no restrictions on weight or computational resources, but the number of observable stars may decrease due to atmospheric conditions, clouds, and ship motion. This study found that increasing the FOV of the measuring equipment can improve the probability of correct matching of star ID in scenarios assuming a maritime environment.
        On the other hand, it should be noted that increasing the FOV may result in lower measurement accuracy, leading to reduced positioning accuracy of the automatic celestial navigation system. Additionally, although not considered in this study, lens distortion increases with a larger FOV. Therefore, if the desired positioning accuracy of the automatic celestial navigation system can be achieved, it is appropriate to increase the FOV of the measuring equipment to improve the probability of correct matching of stars.
    
        This study did not consider lens distortion. Therefore, future studies need to consider the effect of sensor calibration \cite{Klaus2004, Zhang2017, Enright2018} and centroiding techniques \cite{Samaan2002, Liebe2002} for cases with a large FOV. These are considered as future tasks.

    \section{Conclusion}\label{sec:conclude}
        In this study, we investigated the measurement accuracy and probability of correct matching concerning the size of the FOV, focusing on the decrease in the observable magnitude of stars and the presence of stars covered by clouds or the moon in a maritime environment. We discussed the performance specifications required for an automatic celestial navigation system in a maritime environment from the perspective of FOV size. Specifically, assuming the use of CMOS image sensors, commonly used in general cameras for celestial measurements, we calculated the maximum angular resolution for each FOV and its equivalent distance at sea. Additionally, in the investigation of probability of correct matching, we conducted Monte Carlo simulations of star ID to compute the probability of correct matching for each FOV.
    
        The results revealed the following:
        \begin{itemize}
            \item It was found that increasing the FOV can improve the probability of correct matching of star ID using subgraph isomorphism-based methods.
            \item For accurate identification, observing four or more stars is required. However, when the FOV is increased, four stars may be insufficient, at least for $\theta_{\mathrm{FOV}} \geq 80^{\circ}$.
        \end{itemize}
        On the other hand, it should be noted that increasing the FOV may result in lower measurement accuracy, leading to reduced positioning accuracy of the automatic celestial navigation system. Therefore, it is essential to maximize the FOV of the measuring equipment while achieving the desired positioning accuracy for the automatic celestial navigation system in order to improve the probability of correct matching of star ID. 

    \section*{Acknowledgements}
        This research was carried out as part of a collaboration with TOKYO KEIKI INC. and supported by a Grant-in-Aid for Scientific Research from the Japan Society for the Promotion of Science (JSPS KAKENHI Grant \#22H01701 and \#23KJ1432). The authors would like to thank Yuki Tabata, a member of the Ship Intelligentization Sub-Area at Osaka University, for technical discussions. 
    
    \section*{Conflict of interest}
        The study was carried out in cooperation with TOKYO KEIKI INC.
    
    \bibliographystyle{spphys}       
    \bibliography{reference.bib}   
    

\end{document}